\documentclass[letterpaper, 10 pt, conference]{ieeeconf}  

\IEEEoverridecommandlockouts                              

\usepackage{amsmath} 

\usepackage{amsthm}
\usepackage{amssymb}  
\usepackage{cite}
\usepackage{graphicx}
\usepackage{comment}
\usepackage[font=footnotesize]{caption}
\usepackage[subrefformat=parens]{subcaption}
\usepackage{multirow}
\usepackage{booktabs}
\usepackage{bbm}
\usepackage{dsfont}
\usepackage{color}
\let\labelindent\relax
\usepackage{enumitem}
\setlist[itemize]{noitemsep, topsep=0pt, leftmargin=*}
\theoremstyle{definition}
\newtheorem{dfn}{Definition}
\newtheorem{plm}{Problem}

\newcommand{\proposed}{SIAMS}

\title{\LARGE \bf
Reinforcement Learning of Flexible Policies for Symbolic Instructions with Adjustable Mapping Specifications
}

\author{Wataru Hatanaka$^{1}$, Ryota Yamashina$^{1}$, and Takamitsu Matsubara$^{2}$
\thanks{$^{1}$Wataru Hatanaka and Ryota Yamashina are with the Digital Strategy Division, RICOH Company, Ltd, Japan.
        {\tt\small \{wataru.hatanaka, ryohta.yamashina\}@jp.ricoh.com}}%
\thanks{$^{2}$Takamitsu Matsubara is with the Division of Information Science, Graduate School of Science and Technology, Nara Institute of Science and Technology (NAIST), Japan.
        {\tt\small takam-m@is.naist.jp}}%
}

\begin{document}

\maketitle
\thispagestyle{empty}
\pagestyle{empty}

\begin{abstract}
Symbolic task representation is a powerful tool for encoding human instructions and domain knowledge.
Such instructions guide robots to accomplish diverse objectives and meet constraints through reinforcement learning (RL). 
Most existing methods are based on fixed mappings from environmental states to symbols.
However, in inspection tasks, where equipment conditions must be evaluated from multiple perspectives to avoid errors of oversight, robots must fulfill the same symbol from different states.
To help robots respond to flexible symbol mapping, we propose representing symbols and their mapping specifications separately within an RL policy.
This approach imposes on RL policy to learn combinations of symbolic instructions and mapping specifications, requiring an efficient learning framework.
To cope with this issue, we introduce an approach for learning flexible policies called Symbolic Instructions with Adjustable Mapping Specifications (\proposed).
This paper represents symbolic instructions using linear temporal logic (LTL), a formal language that can be easily integrated into RL.
Our method addresses the diversified completion patterns of instructions by (1) a specification-aware state modulation, which embeds differences in mapping specifications in state features, and (2) a symbol-number-based task curriculum, which gradually provides tasks according to the learning’s progress.
Evaluations in 3D simulations with discrete and continuous action spaces demonstrate that our method outperforms context-aware multi-task RL comparisons.
\end{abstract}

\vspace{-3mm}
\section{Introduction}
Performing tasks accurately according to given instructions is an essential function of robots, and reinforcement learning (RL) is one promising approach through which they can learn this skill \cite{kober2013reinforcement}.
Symbolic representations have played an important role in task modeling for RL \cite{brachman2004knowledge}.
This approach enables the intuitive implementation of prior domain knowledge and serves as high-level plans for RL by formal languages \cite{lyu2019sdrl, illanes2020symbolic, jin2022creativity}.
The strict description rules of formal language act as a practical guide for sequential decision-making in RL, allowing efficient learning of explainable and generalizable policies \cite{glanois2024survey}.
\begin{figure}[t]
    \centering
    \includegraphics[width=86mm]{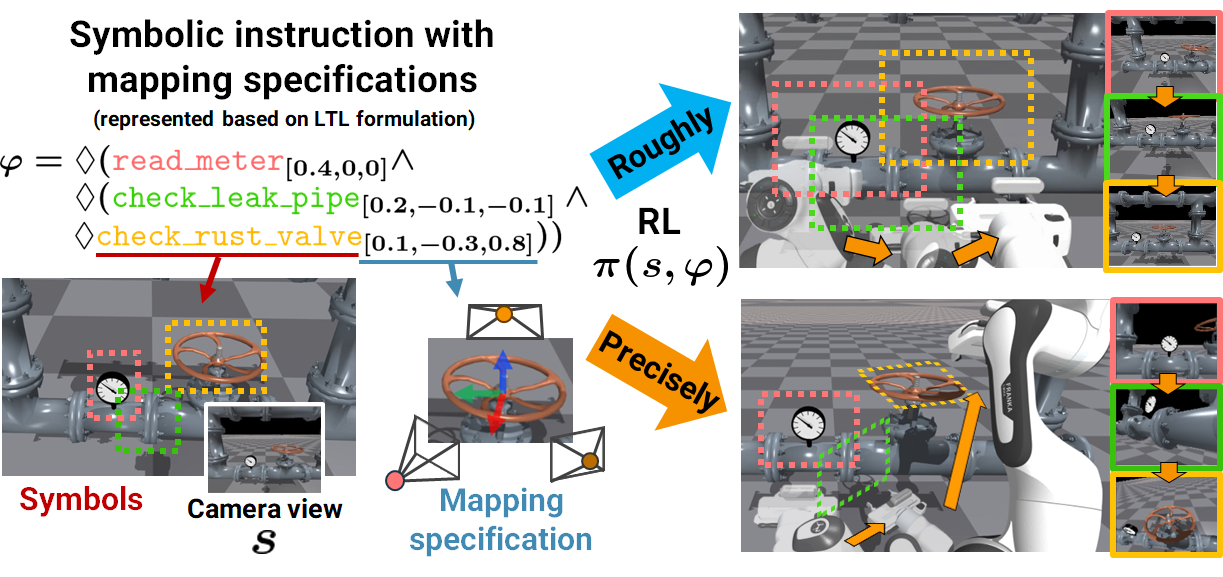}
    \caption{Overview of our method: By introducing a symbol mapping specification, a policy learns to follow symbol instructions and satisfaction criteria for each symbol. Our method allows robots to perform such sequential tasks as inspections that require checking from different perspectives based on requested level of accuracy.}
    \label{fig:overview_intro}
    \vspace{-3mm}
\end{figure}

Grounding the events occurring in environments to symbols is essential for an RL policy led by symbolic instructions for adapting to the real world.
In particular, the design of the mapping specification, which specifies the environmental states to be mapped to symbols, such as the robot's position and pose, affects their performance in the logic-based models \cite{ghallab1998pddl, Pnueli1977-ug}.
This mapping specification has been fixed to allow consistent learning of symbolic instructions in RL.

However, the challenges posed by this inflexibility in real-world applications of robots have been overlooked.
For example, some tasks require inspection of identical equipment from two different perspectives concerning accuracy levels: roughly during routine checks and precisely during emergencies (Fig. \ref{fig:overview_intro}).
This situation requires a robot to fulfill symbols like ``$\mathtt{inspect\_piping}$'' from different perspectives.
If mapping specifications are fixed during training, the policy must be relearned whenever these specifications change. 
Representing a robot's perspective with symbols can be handled within conventional frameworks using modifiers, e.g., ``$\mathtt{inspect\_piping\_from\_nearby}$''; however, this situation requires preparing symbols that account for every possible pose the robot might encounter, which is quite impractical. 
Therefore, symbol mappings should be \textit{adjustable} based on mapping specifications so that the robot can fulfill each symbol under different conditions.
Nevertheless, mapping specifications complicate the symbol's satisfaction criteria and exponentially diversify the completion pattern per instructions, making an efficient framework essential.

To address this issue, we propose an approach with symbolic instructions with adjustable mapping specifications (\proposed), which allows the robot to follow diverse symbolic instructions and their completion conditions through RL.
In this paper, we present an implementation of \proposed{} that learns a policy that follows linear temporal logic (LTL) instructions with symbol mapping explicitly conditioned by mapping specifications.
LTL can be integrated into off-the-shelf RL algorithms by making its completion an optimization objective through reward functions \cite{toro2018teaching, icarte2018using, camacho2019ltl, Vaezipoor2021-qm}.
We improve the efficiency of learning diversified task completion patterns by introducing two key technical components: specification-aware state modulation and symbol-number-based task curriculum.
The former learns the embeddings of environmental states based on mapping specifications through RL, allowing a policy to follow the specifications without additional dimensions in the parameter space.
The latter, which is based on the number of symbols within the LTL instructions, automatically provides tasks with increasing difficulty to the policy based on the learning progress.
Our method is evaluated on robot simulations in a 3D navigation task with discrete action space and inspection tasks with continuous action space under image input and sparse reward.
Experimental results show that our method outperforms other context-aware multi-task RL methods.

Our contributions are as follows:
\begin{itemize}
    \item Proposal of \proposed{}, an approach that learns to accomplish symbolic instructions with adjustable symbol mapping specifications through RL;
    \item Development of a learning framework for diverse LTL instructions with adjustable symbol mapping, integrating specification-aware state modulation and a symbol-number-based task curriculum;
    \item Empirical evaluations in 3D simulation with image input; a navigation task in discrete action spaces and a robotic inspection task in continuous action spaces.
\end{itemize}

\section{Related works}
\subsection{Mapping Design in Symbolic Instruction-guided RL}
Symbolic instructions have been integrated into RL so that agents can learn to perform sequential decision-making through various formal methods, such as automata \cite{icarte2018using, camacho2019ltl}, action-based models \cite{lyu2019sdrl, illanes2020symbolic, jin2022creativity}, and logic-based models \cite{toro2018teaching, Vaezipoor2021-qm, hatanaka2023reinforcement}.
These methods have mainly been evaluated only in toy environments like grid worlds, where the locations of symbolic events are fixed and the agent can easily identify them from the state.
Specifying Tasks for Reinforcement Learning (SPECTRL) \cite{jothimurugan2019composable} provides a specification language that can indicate the condition of the symbol mapping by augmenting temporal logic.
Although this method allows reward shaping in RL by mapping specifications, it requires relearning when they change.
Context-aware probabilistic temporal logic (CAPTL) \cite{CAPTL2020} also proposes a formal language for context-aware symbolic task transitions, but it only proposes a way to model a system.

Unlike these methods, we propose an RL framework that adjusts the conditions under which each symbolic event is fulfilled based on the mapping specifications while following symbolic instructions.

\subsection{Context-aware Multi-task RL}
Task metadata such as mapping specifications are used as a context for conditioning task-agnostic policy learning in multi-task RL literature.
Although using predefined IDs that uniquely identify tasks as context is a traditional approach \cite{yu2020meta, yang2020multi, sodhani2021multi}, methods have been proposed in recent years that use the natural language embedding of task descriptions or instructions for learning \cite{shridhar2023perceiver}.

These approaches are related to our method, which aims to learn policy to accomplish tasks in diverse contexts.
However, our method can incorporate domain knowledge through symbolic instructions while allowing the achievements of each symbol to be adjustable by continuous values based on mapping specifications.
Such flexibility distinguishes our method from previous works.

\section{Preliminaries}
\subsection{Symbolic Instructions by LTL}
We use LTL to formally represent symbolic instructions.
An LTL formula consists of a finite number of propositional symbols $\mathcal{PS}$, logical operators to connect propositions such as $\land$ (conjunction), $\lor$ (disjunction), and $\lnot$ (negation), and temporal operators to represent the temporal relations of propositions, such as $\Diamond$ (eventually), $\cup$ (until), and $\bigcirc$ (next).
\noindent
\textbf{Example of task description by LTL.}
The inspection task ``read the meter and check for rust on the valve'' can be described as $\varphi=\Diamond (\mathtt{read\_meter}\land \Diamond \mathtt{check\_rust\_valve} )$.
More details of LTL syntax can be found in a previous work \cite{baier2008principles}.
For episodic tasks in RL, we treat co-safe LTL (sc-LTL) \cite{kupferman2001model, lacerda2015optimal}, which is a subclass of LTL that deals with formulas that can be verified within a finite number of steps.
Henceforth, we denote sc-LTL as LTL in this paper.

\subsection{Reinforcement Learning for LTL Objectives}
The interaction of an agent with its environment is commonly modeled by a Markov Decision Process (MDP) in RL and defined by the following tuple, $\mathcal{M}=( S, T, A, P, R, \gamma, \mu)$, where $S$ is a set of states, $T\subseteq S$ is a set of terminal states, $A$ is a set of actions,  $P : S \times A \times S \rightarrow [0,1]$ is a state transition function, $R : S \times A \rightarrow \mathbb{R}$ is a reward function, $\gamma \in(0,1)$ is a discount factor, and $\mu : S\to[0,1]$ is a distribution over the initial states.

For an agent to learn behavior to complete LTL tasks, it must know the current task's progress and the next symbol to be fulfilled from it.
Symbol mapping function $E:S\rightarrow \{0,1\}^{\mathcal{PS}}$ is used for mapping the current environmental state to pre-defined symbols.
This paper describes \textit{satisfy} as the agent achieving the state that the function $E$ maps to any symbol $p\in\mathcal{PS}$, which is necessary for linking LTL task transition and MDP.
The satisfied symbol is verified against the given LTL formula $\varphi$ by a progress function $\operatorname{prog}(E(s), \varphi)$, which manages task improvement by implementing the progression algorithm \cite{bacchus2000using}.

\noindent
\textbf{Example of LTL progression.}
The inspection task $\varphi=\Diamond (\mathtt{read\_meter}\land \Diamond \mathtt{check\_rust\_valve})$ is rewritten by leaving symbols unsatisfied to $\varphi'=\operatorname{prog}(E(s), \varphi)=\Diamond \mathtt{check\_rust\_valve}$ when the symbol $\mathtt{read\_meter}$ is satisfied in state $s$.

The function returns $\mathsf{true}$ if the LTL formula completes, $\mathsf{false}$ if it is violated, and the original LTL otherwise.
We model an RL agent with an LTL objective using Taskable MDP \cite{Vaezipoor2021-qm}, which exploits the progress function to define a Markovian reward by synchronizing the LTL task transitions with the environmental state transitions in the MDP.
A Taskable MDP is defined as follows.
\begin{dfn}[\textit{Taskable MDP}]
  Given an MDP without a reward function as tuple $\mathcal{M}=\left( S, T, A, P, \gamma, \mu \right)$, finite set of symbols $\mathcal{PS}$, symbol mapping function $E$, finite set of LTL formulas $\Phi$ and its probability distribution $\tau_{\Phi}$ over $\Phi$, a Taskable MDP is defined as tuple $\mathcal{M}_{\Phi}=\left( S^{\prime}, T^{\prime}, A, P^{\prime}, R^{\prime}, \gamma, \mu^{\prime}\right)$,  where $S^{\prime}=S\times cl(\Phi)$ is a finite set of product states and $cl(\Phi)$ (where $cl$ stands for closure) is the smallest set containing $\Phi$, including every formula $\varphi\in\Phi$ and all formulas derived through its progression.
  $T'=\{\langle s,\varphi\rangle|s\in T \ \text{or} \  \varphi\in\{\mathsf{true, false}\}\}$ is a terminal set of product states, and $P^{\prime}$ is the transition probability of the product states; $P^{\prime}\left(\langle s^{\prime}, \varphi^{\prime}\rangle \mid\langle s, \varphi\rangle, a\right)=P\left(s^{\prime} \mid s, a\right)$ if $\varphi^{\prime}=\operatorname{prog}(E(s), \varphi)$ otherwise zero, $\mu^{\prime}(\langle s, \varphi\rangle)=\mu(s) \cdot \tau_{\Phi}(\varphi)$ is an initial distribution of product states and reward function 
\begin{equation}
R^{\prime}(\langle s, \varphi\rangle, a)= \begin{cases}1 & \text { if } \operatorname{prog}(E(s), \varphi)=\mathsf { true } \\ -1 & \text { if } \operatorname{prog}(E(s), \varphi)=\mathsf { false } \\ 0 & \text { otherwise }\end{cases}.
\label{eq:reward_ltl2action}
\end{equation}
\end{dfn}
Set $\Phi$ contains LTL formulas consisting of combinations of pre-defined symbols and LTL operators, and policy $\pi: S'\times A\rightarrow [0,1]$ is optimized to accomplish all the LTL formulas sampled from set $\Phi$ through RL by maximizing reward $R'$ for every state $s\in S$.

\section{Problem formulation}
To enable adjustments in how the policy satisfies each symbol by changing the mapping specifications, we implement this by augmenting the capabilities of the symbol mapping function.
\subsection{Specification-Aware Symbol Mapping}
\begin{figure}[t]
    \centering
    \includegraphics[width=84mm]{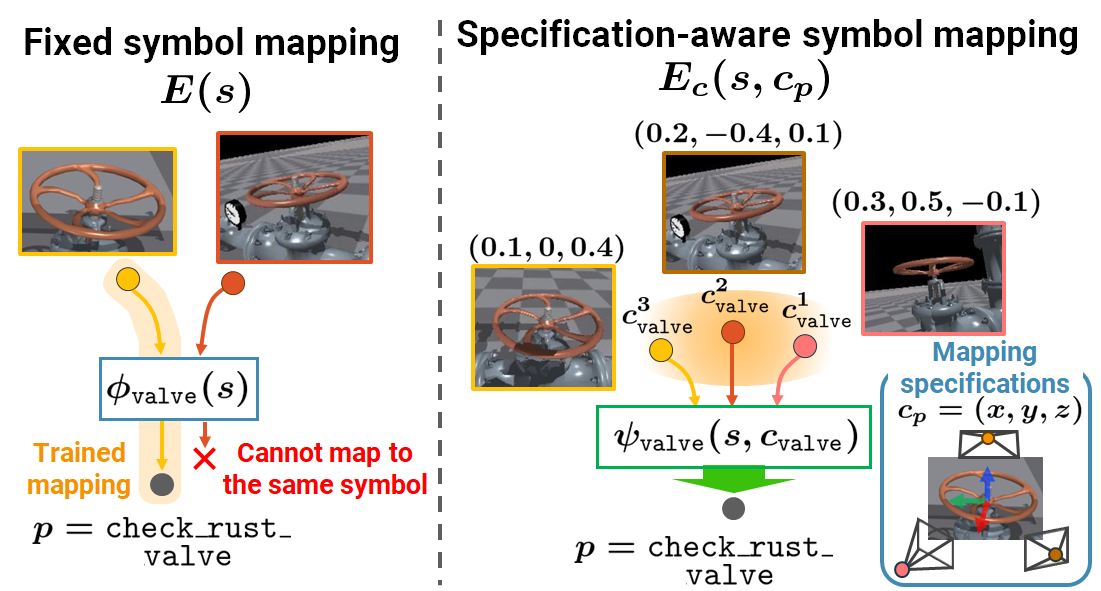}
    \caption{Comparison of specification-aware symbol mapping with fixed mapping: Our symbol mapping allows different states to satisfy identical symbols based on instructed mapping specifications.}
    \label{fig:specification_mapping}
    \vspace{-2mm}
\end{figure}

First, we introduce an \textit{evaluation function} $\phi_p:S\rightarrow \{0,1\}$ that defines a state in which each symbol $p\in \mathcal{PS}$ is satisfied.
We can rewrite symbol mapping function $E$:
\begin{equation}
    E(s)=\{p\in\mathcal{PS}\mid \phi_p(s)=\mathsf{true}\}.\nonumber
\end{equation}
\noindent
The function $\phi_p$ is typically designed using threshold processing based on mapping specifications, such as an agent's position or pose.
Existing works condition these specifications solely on state $s$, assuming they remain fixed during training.
However, altering mapping specifications post-training requires policy retraining.
To overcome this limitation, we treat specifications as adjustable parameters during training and formulate the mapping function as follows:
\begin{equation}
    E_c(s,c_p)=\{p\in\mathcal{PS}\mid \psi_p(s,c_p)=\mathsf{true},c_p\in C_p\},
\end{equation}
where $\psi_p:S\times C_p\rightarrow \{0,1\}$ is the evaluation function explicitly conditioned by mapping specification $c_p\in C_p$.
We call this function a \textit{specification-aware symbol mapping function} and define it as follows.
\begin{dfn}
 Given set of states $S$, finite set of symbols $\mathcal{PS}=\{p_1,p_2\cdots,p_n\}$, and evaluation function $\psi_p$, a specification-aware symbol mapping is defined as a function, $E_c:S\times \mathcal{C}\rightarrow 2^{\mathcal{PS}}$, where $\mathcal{C}=\{c_{p_1},c_{p_2},\cdots,c_{p_n}\}$ is a finite set of the mapping specifications for all symbols $\forall p\in\mathcal{PS}$.
\end{dfn}
We show the difference between specification-aware symbol mapping and traditional fixed mapping in Fig. \ref{fig:specification_mapping}.
When we use camera viewpoints as mapping specifications, the specification-aware symbol mapping allows us to train flexible inspection policy by adjusting satisfaction criteria for each symbol according to given specifications without policy retraining.
If the finite set of mapping specifications $\mathcal{C}$ is always fixed, $E_c$ is equivalent to existing symbol mapping $E$.
Specification-aware symbol mapping function $E_c$ can be integrated with the Taskable MDP by replacing symbol mapping $E$ with $E_c$ and state transition $P'$ with $P_c\left(\langle s^{\prime}, \varphi^{\prime}\rangle \mid\langle s, \varphi, \mathcal{C}\rangle, a\right)$, where $\varphi^\prime=\operatorname{prog}(E_c(s,\mathcal{C}),\varphi)$.

\noindent
\textbf{Example design of mapping specifications.}
If set of states $S$ is defined as an agent position in a two-dimensional Euclidean space $(x,y)\in\mathbb{R}^2$, satisfaction of symbol ``$\mathtt{read\_meter}$'' is determined from the fact that the agent has captured a meter from which the distance $dist(x,y)$ is less than a threshold value: $\mathcal{C}=\{c_{p_1}\mid c_{p_1}\in[0,D],D\in\mathbb{R}\}$ and $\psi_p(\langle x,y\rangle,c_{p_{1}})=\mathsf{true}$ if $dist(x,y)\leq c_{p_{1}}$ else $\mathsf{false}$.
To facilitate quantitative evaluation in experiments, this paper assumes that mapping specification $c_p$ is sampled according to probability distribution $\tau_{c_p}$ over finite set $C_p$, and evaluation function $\psi_p$ is defined in a form that specifies directly the constraints of the state space.
\subsection{Problem Statement}
This paper aims to obtain optimal policy $\pi$ for $\mathcal{M}_{\Phi}$ that maximizes reward by accomplishing any LTL tasks $\varphi\in\Phi$ and any set of mapping specifications $C_p$ under specification-aware symbol mapping function $E_c$.
Now the problem can be formally presented.
\begin{plm}
Given a Taskable MDP without product state transition $\mathcal{M}_{\Phi}=\left( S', T', A, R', \gamma, \mu'\right)$, finite set of symbols $\mathcal{PS}$, a finite set of mapping specifications for each symbol $C_p$, probability distribution $\tau_{c_p}$ over $C_p$, specification-aware symbol mapping function $E_c$, finite set of LTL formulas $\Phi$, and probability distribution $\tau_{\Phi}$ over $\Phi$, our goal is to find policy $\pi$ that maximizes reward $R'$ under state transition $P_c$ for every state $s\in S$, LTL formula $\varphi\in\Phi$, and a finite set of mapping specifications for all symbols $\mathcal{C}=\{c_p\mid c_p\in C_p,\forall p\in\mathcal{PS}\}$.
\end{plm}

\begin{figure*}[t]
    \centering
    \includegraphics[width=176mm]{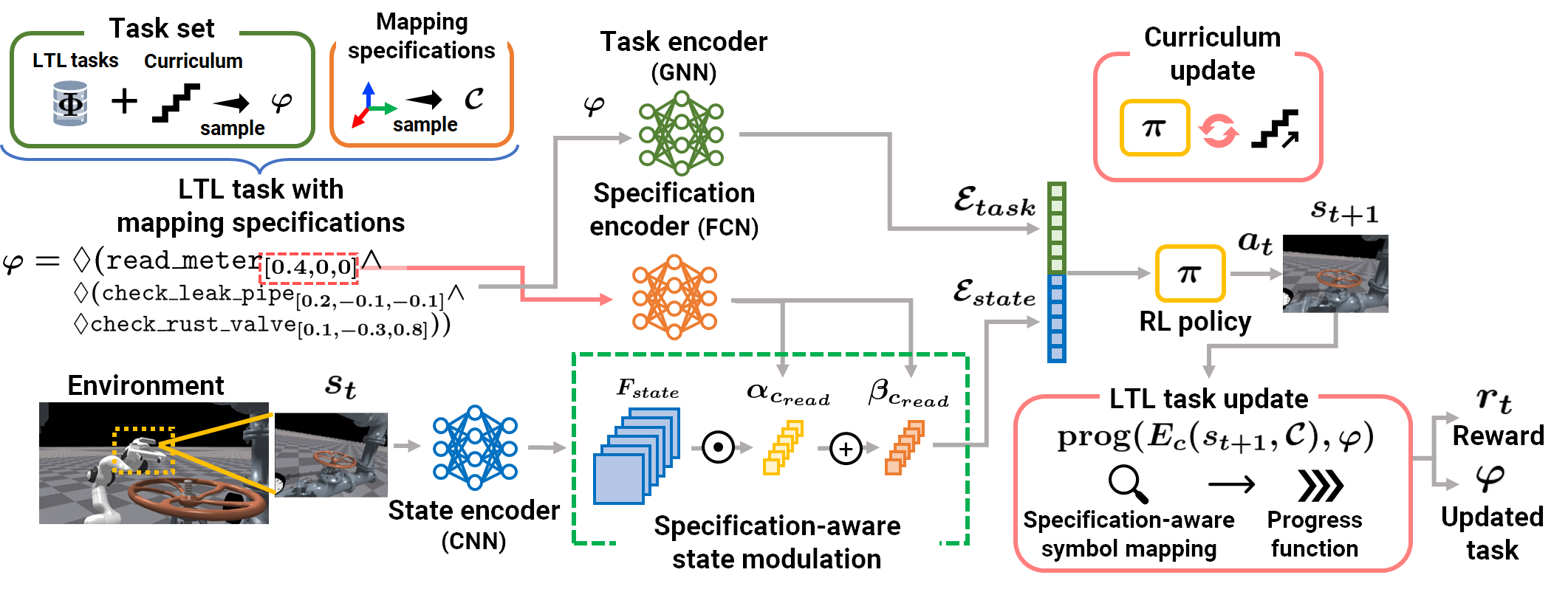}
    \caption{Overview of \proposed{} framework: For simplicity, we describe LTL and mapping specifications as LTL tasks with mapping specifications based on LTL’s formulation. When next symbol of a given LTL task is $\mathtt{read\_meter}$, state embedding $\mathcal{E}_{state}$ is modulated by affine parameters, $\alpha_{c_{read}}$ and $\beta_{c_{read}}$, conditioned by mapping specification $c_{read}$, and is passed to the policy along with task embedding $\mathcal{E}_{task}$. All encoders and policy $\pi$ are optimized through RL by reward defined in Eq. \ref{eq:reward_ltl2action}.}
    \label{fig:overview_proposed}
\end{figure*}

\section{Proposed framework}
\subsection{Overview of Training Framework of \proposed}
Fig. \ref{fig:overview_proposed} shows an overview of our proposed framework.
In the training phase, LTL task $\varphi$ is uniformly sampled from set $\Phi$ based on the curriculum described in Section \ref{LTL_curriculum} at the beginning of the episode.
Similarly, mapping specification $c_p$ is uniformly sampled from each set $C_p$, forming a set of mapping specifications for all symbols $\mathcal{C}$.
A given LTL task is input to the policy through the task encoder.
To accomplish diverse LTL tasks in a sparse reward environment, it is necessary for a policy to learn actions for the entire LTL, not just each symbol.
Therefore, we use a graph neural network as the task encoder that learns the embeddings of LTL formulas transformed into a graph \cite{Vaezipoor2021-qm}.
The specification encoder learns features for modulating the state embedding described in Section \ref{FiLM}.
Mapping specifications are individually designed for each symbol; however, the specifications of symbols to be satisfied in the future have minimal impact on the policy's actions to accomplish the next symbol. To simplify the learning process, we input only the next symbol's specification to the specification encoder.
$E_c$ receives next state $s_{t+1}$ reflecting the action from policy $a_t$, and reward $r_t$ is calculated by Eq. \ref{eq:reward_ltl2action}, and the LTL task is updated by the progress function $\varphi:=\operatorname{prog}(E_c(s_{t+1},\mathcal{C}),\varphi)$.

\subsection{Specification-aware State Modulation}\label{FiLM}
Most context-aware multi-task RL methods seek a common policy for different tasks under state space $S^C$ explicitly conditioned by task context $C$ \cite{du2019provably, yu2020meta, yang2020multi, sodhani2021multi}.
Unlike these approaches, we assume that the output of evaluation function $\psi$ for each symbol $p\in\mathcal{PS}$ varies across mapping specification $c_p$, even for identical state $s\in S$.
For the policy to respond to differences in mapping specifications, it must capture the environment state space constrained by specification-aware symbol mapping $E_c$; the integrated representation of the environmental state, the LTL task, and the mapping specifications must be discriminable for the policy in the feature space.
A naive approach is to concatenate each feature; however, this choice increases the dimensions of the policy input, complicating learning.

To address this issue, we propose a method for learning the embeddings of the environmental state according to symbol mapping specifications.
This strategy is based on the perspective that the symbols dominantly determine the states the policy should visit, and their mapping specifications can be reasonably represented by modulating the state sequences to satisfy them.
We implement this approach incorporating FiLM \cite{perez2018film}, which learns affine parameters that modulate the encoded intermediate features according to the context.
Mapping specification $c_p$ is used as the conditioning context and modulates the features of state encoder $F_{state}$:
\begin{eqnarray}
    \mathcal{E}^{c}_{state} = \alpha_{c_p} F_{state} + \beta_{c_p}, \\
    \text{where}\ \alpha_{c_p}, \beta_{c_p}=f_{c}(c_p),\ c_p\in \mathcal{C}. \nonumber
\end{eqnarray}
\noindent
$\alpha_{c_p}$ and $\beta_{c_p}$ are affine parameters for the modulation.
In this paper, ${F}_{state}$ is a feature map of a convolutional neural network (CNN), and specification encoder $f_{c}$ consists of a 1-layer fully connected network in which network parameters are shared for generating each affine parameter.

\subsection{Symbol-number-based Task Curriculum}\label{LTL_curriculum}
The variety of symbol instructions that the agent can learn depends on LTL task set $\Phi$, and as the number of symbols increases within a single task, the steps required to accomplish them also increase.
Besides, the interaction increases with the environment for policy learning as the set of mapping specifications grows, and the reward that can only be given after completing the LTL becomes sparser, making learning convergence more complicated.

To address this, we define difficulty levels based on the number of symbols in the LTL and gradually introduce them via a curriculum as learning progresses.
During training, we periodically test the policy on each available LTL task with a different seed and use the average success rate to determine when to advance to the next level.
To avoid performance degradation on experienced tasks, each episode’s LTL task is uniformly sampled from all tasks at or below the current level.
This allows the policy to gradually tackle complex LTL tasks as learning progresses.
\section{Experiments}
We empirically evaluated our method in a challenging environment in which a policy is given different LTL and mapping specifications for each episode.
We conducted two visual inspection tasks: discrete-action navigation in a 2D environment for scalability with respect to the diversity of LTL and mapping specifications, and continuous-action robotic arm control in a 3D environment for realistic applicability.
\subsection{Comparisons}
To clarify our method's performance, we compared it from two perspectives: a task conditioning approach and a curriculum strategy.
\subsubsection{Task conditioning approach}
We compare our method to Soft-Module\cite{yang2020multi} and CARE\cite{sodhani2021multi}, state-of-the-art architectures baseline for learning task embedding from continuous value inputs in multi-task RL, and Naive, an ablation of the specification-aware state modulation from ours.  
\begin{itemize}
    \item \textbf{Soft-Module:} This method trains routing networks that weigh connections between the state encoder’s layers using context vectors. A flattened CNN feature is input to the state encoder, and a concatenated task and specification vector are input to the routing network. Following the original paper, the state encoder has two layers and two modules per layer. We omit loss weighting by task difficulty, as the task curriculum serves the same purpose.
    \item \textbf{CARE:} This method trains multiple state encoders and a context encoder to compute attention scores for mixing their features. Following the paper, we train six CNNs, flatten their outputs, and then mix them. A concatenated one-hot task and a specification vector are encoded with a one-layer fully connected network. Finally, the state and context encodings are concatenated and given to the policy.
    \item \textbf{Naive:} This method encodes a one-hot task vector and a specification vector through a one-layer fully connected network. The encoded features are then concatenated with the flattened CNN features and input into the policy.
\end{itemize}

\subsubsection{Curriculum strategy}
\begin{itemize}
    \item \textbf{Anti-curriculum:} This method reverses the curriculum order, known as ``hard to easy'' \cite{2022surveycurriculum}, and places the LTL task with the most symbols at the first level.
    \item \textbf{No curriculum:} All the levels of the tasks are registered in the task set from the start of training, and this ablation does not apply a curriculum to our method.
\end{itemize}

\subsection{Training Settings}
We implement our method on top of LTL2Action \cite{Vaezipoor2021-qm}\footnote{https://github.com/LTL2Action/LTL2Action}.
All baselines, including ours, use PPO \cite{schulman2017proximal} for training.
Each actor and critic is a 3-layer fully connected network.
To encourage efficient learning, a negative reward of -0.01 per step is given in all the experiments.
We used $80\times60$ RGB images as input, and the state encoder is a 3-layer CNN with ReLU.
The task encoder is a graph neural network identical to LTL2Action, with the same hyperparameters.

This paper assumes that the specification-aware symbol mapping and the mapping specification are provided manually and accurately.
At the start of each episode, we uniformly sample an LTL task $\varphi$ and a set of symbol mapping specifications $\mathcal{C}$.
We test the policy with 20 episodes every 25 updates to determine the curriculum level.

\begin{figure}[t]
    \centering
    \includegraphics[width=85mm]{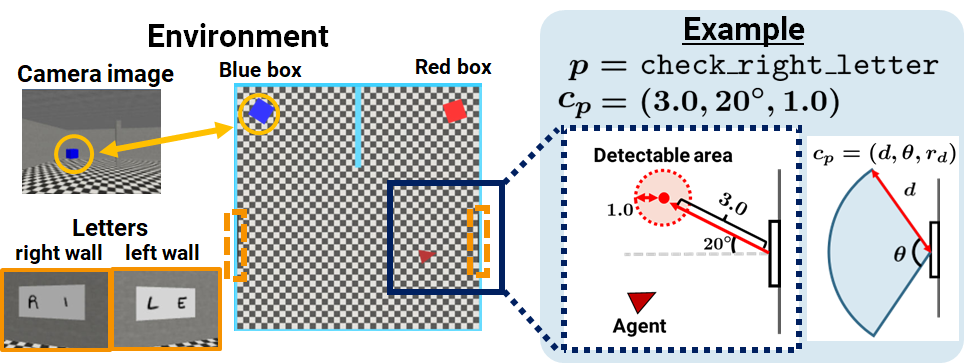}
    \caption{Environment of navigation task: Map has red and blue boxes and letter ``LE'' on left wall and ``RI'' on right one. Symbol mapping specifications for letter symbols are defined by distance $d$ and angle $\theta$ from position of letters and radius $r_d$ of detectable area.}
    \label{fig:navigation_environments}
    \vspace{-2mm}
\end{figure}
\subsection{Navigation Task with Discrete Action Space}\label{navigation_task}
We built a 2D environment for visual inspection using Miniworld \cite{MinigridMiniworld23}.
Fig. \ref{fig:navigation_environments} shows a bird’s-eye view and an agent-mounted camera image.
The $10 \text{m} \times 10 \text{m}$ map has two differently colored boxes, and the agent starts near the bottom center with a random pose.
Its actions are \{\textit{turn-left}, \textit{turn-right}, \textit{move-forward}\}, where \textit{move-forward} advances $0.5 \text{m}$ and \textit{turns} rotate the agent by $30$ degrees.
Each episode lasts up to $150$ steps, after which it ends with zero reward.
All methods were trained for $40$ million steps in Scenario 1 and $50$ million steps in Scenario 2.
\subsubsection{Symbols and mapping specifications}
We defined four symbols: \texttt{reach\_bluebox}, \texttt{reach\_redbox}, \texttt{check\_letter\_left} and \texttt{check\_letter\_right}. The \texttt{reach} symbols are satisfied when the agent reaches the corresponding box, while \texttt{check\_letter} symbols are satisfied when the agent sees the letter from the position specified by the mapping.
Visual examples of the \texttt{check\_letter} symbols and their mapping specifications are provided in Fig. \ref{fig:navigation_environments}.
The evaluation function $\psi_p$ outputs $\mathsf{true}$ if the agent meets two conditions: 1) entering the detectable area defined by the mapping specification, and 2) capturing the letter in view. 
The mapping specification defines the detectable area's location as radius $r_d$, with distance $d$ and angle $\theta$ from the xy coordinate centered on each letter.
Here, we fixed $r_d=1.0$ and specifications as $c_p=(d,\theta,r_d)$ where $d\in[1.0,4.0], \theta\in[-60^\circ,60^\circ]$. 
While using image recognition is realistic, we used a minimal setup to focus on policy learning performance.
The second condition is simplified: it is satisfied if the angular difference between the agent's heading and the $(d,\theta)$ vector is within $20$ degrees.

\subsubsection{Task and curriculum settings}
We evaluated the performance of our method through two sequential tasks of different difficulties: Scenario 1 and Scenario 2.
Scenario 1 has four symbols, and an agent is tasked with checking each letter once within an episode: $\mathcal{PS}_1=\{p_1, p_2, p_3, p_4\}$, where $p_1=\texttt{reach\_bluebox}$, $p_2=\texttt{reach\_redbox}$, $p_3=\texttt{check\_letter\_left}$, $p_4=\texttt{check\_letter\_right}$.
Scenario 2 has six symbols and requires an agent to check each letter twice with different mapping specifications: $\mathcal{PS}_2=\{p_1, p_2, p_3, p_3, p_4, p_4\}$.
Each scenario has levels corresponding to the number of symbols, i.e., Scenario 1 has four, and Scenario 2 has six. 
At each level, the task set is composed of a combination of LTL formulas connected by $\Diamond$ (eventually) operators where the number of symbols equals the level.
The level is initially set to one.
It increases by one if the average success rate of all the tasks, below the current level by the policy rollout conducted during training at regular intervals, exceeds $0.9$ on average.

\subsubsection{Results}
\begin{figure}[t]
    \centering
    \includegraphics[width=88mm]{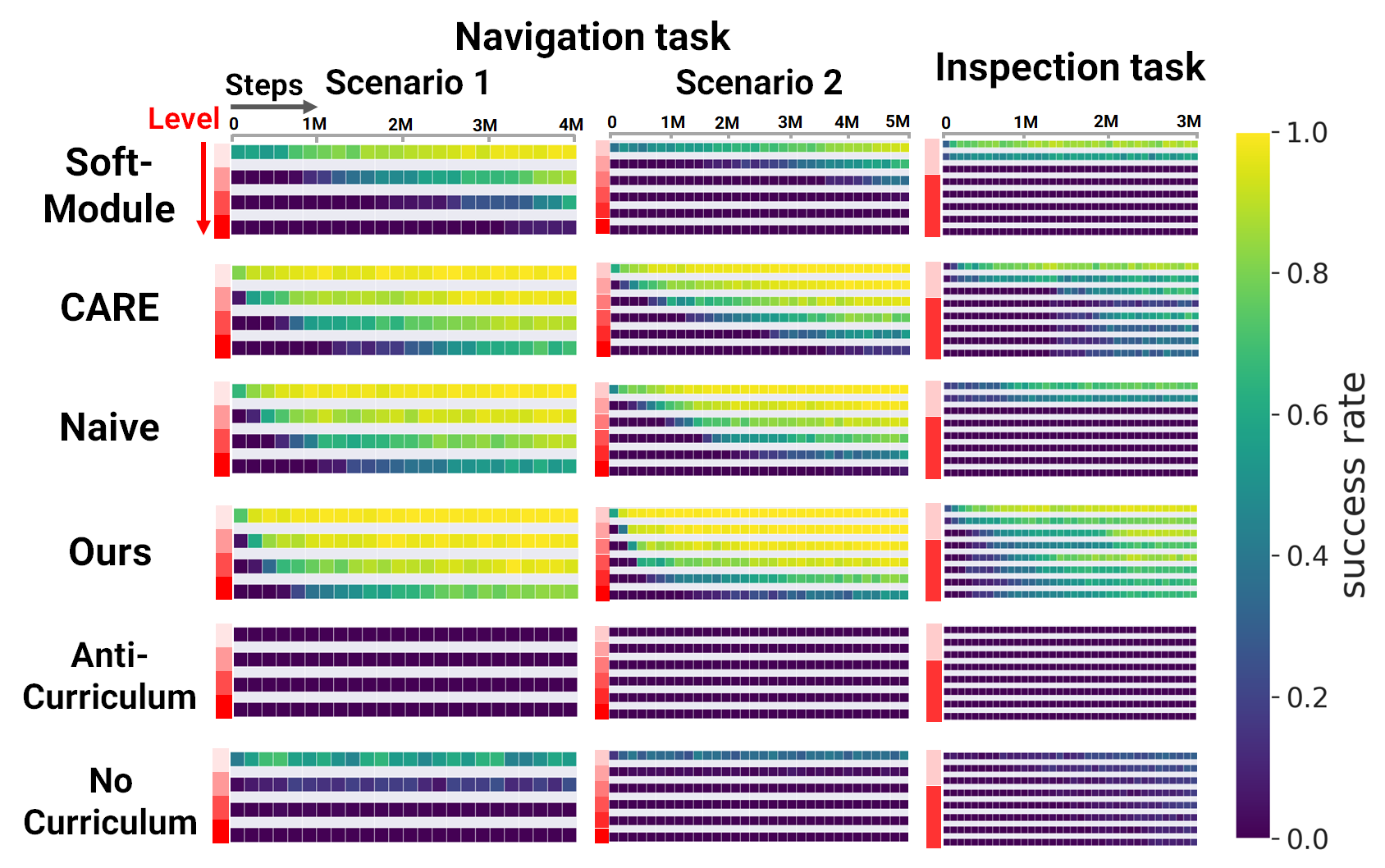}
    \caption{Evaluation results: They are averaged over three random seeds for both navigation and robotic inspection tasks. Each plot is arranged by curriculum level, and the x-axis shows the number of steps. Brightness of square color indicates the average success rate at that step. We periodically evaluate the training policy using a different seed than training, running 20 episodes for each task available at the current level, and report the average success rate.}
    \label{fig:result_all}
    \vspace{-3mm}
\end{figure}

\textbf{Performance evaluation against context-aware RL methods.}
Fig. \ref{fig:result_all} shows the performance in each scenario and level throughout the training.
The brightness of the block colors indicates the average success rate for each task in the policy test during training.
Our method most quickly transitions to brighter colors as the learning steps increase.
This indicates that in both scenarios, it reached the highest curriculum level the fastest and effectively proceeds with learning even with long-horizon tasks.
Table \ref{tab:result} summarizes the quantitative evaluation results.
The performance of the conventional method in Scenario 2 deteriorates significantly, indicating the challenges of accomplishing LTL diversified by mapping specifications.
In contrast, our proposed method maintained superior performance, demonstrating its necessity for addressing this challenge. 

CARE performed second best throughout experiments, whereas Soft-Module performed worse than the Naive one.
These results are consistent with our perspective that differences in mapping specifications should be efficiently conditioned within the feature space of environmental states.
\begin{figure}[t]
    \centering
    \includegraphics[width=88mm]{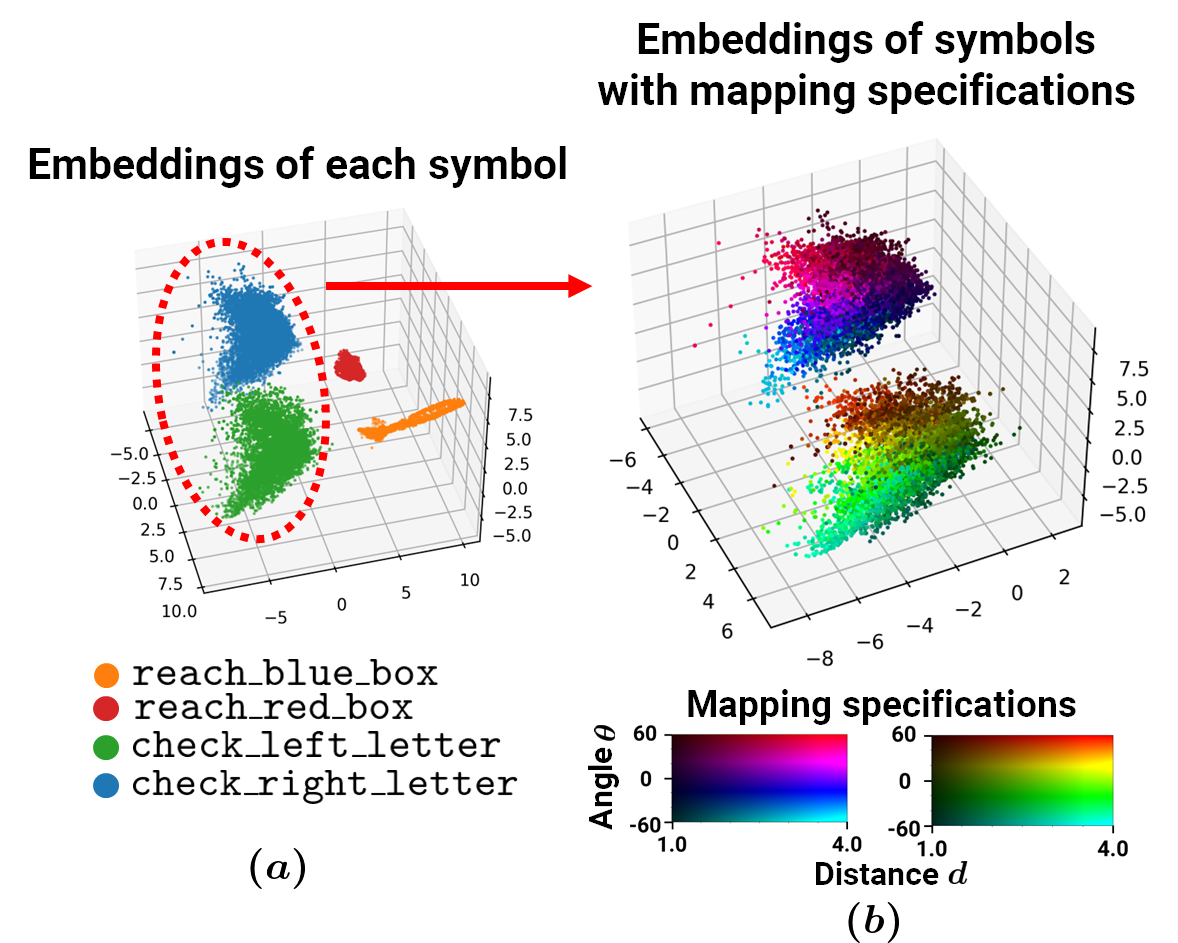}
    \caption{(a) Visualization of embeddings from 1000-episode rollouts: Embeddings are projected into 3D space using PCA, and colors represent the next symbol for each episode. (b) Visualization of mapping specification embeddings: Specifications are defined by the angle $\theta$ and distance $d$ relative to each letter’s detectable area. Color and brightness indicate specification variations.}
    \label{fig:embedding}
\vspace{-3mm}
\end{figure}
\noindent
\textbf{Performance evaluation against curriculum strategy.}
In our experiments, the performance variation among different curricula is significant, where both Anti-curriculum and No-curriculum have greatly degraded performances.
The main cause of their learning collapse is that the probability of sampling easy tasks with a small number of symbols becomes relatively low in the early stage of the training, making the reward too sparse to learn.
This result indicates that a step-by-step strategy for learning based on difficulty is essential, and a symbol-number-based task curriculum can be useful.

\noindent
\textbf{Visualization of learned features by \proposed.}
We also verified that our method can learn discriminative features according to the symbols in LTL tasks and their mapping specifications through RL.
Fig. \ref{fig:embedding}(a) shows the concatenated features of $\mathcal{E}_{state}$ and $\mathcal{E}_{task}$, obtained through the rollout of the trained policy in Scenario 2, by mapping them to a three-dimensional vector by principal component analysis (PCA).
Each data point in the plot corresponds to a feature at a certain step, and each color corresponds to the next symbol given to the agent.
The features are mapped to a separate space for each symbol.
The spread of the green and blue clusters of symbols $\mathtt{check\_left\_letter}$ and $\mathtt{check\_right\_letter}$ intuitively represents the size of the mapping specifications indicated by distance and angle, compared to the orange and red clusters corresponding to symbols $\mathtt{reach\_blue\_box}$ and $\mathtt{reach\_red\_box}$.

To observe the differences in mapping specifications represented in the feature space in more detail, we visualized the features focused on symbols with mapping specifications (Fig. \ref{fig:embedding}(b)).
Each cluster is colored based on the values of each mapping specification, $d$ and $\theta$.
The results show that each symbol and its mapping specification changes are represented continuously and intuitively in the feature space.
This indicates that our method can learn discriminative features that effectively capture the differences between symbols and their mapping specifications in LTL tasks.

\subsection{Inspection Task with Continuous Action Space}\label{inspection_task}
We built a 3D piping inspection task using a robotic arm-mounted camera in Isaac Gym \cite{makoviychuk2021isaac} (Fig. \ref{fig:inspection_ex}(a)).
\subsubsection{Symbols and mapping specifications}
We defined three symbols, $\mathtt{read\_meter}$, $\mathtt{check\_rust\_valve}$, and $\mathtt{check\_leap\_pipe}$, and constructed inspection tasks from their mapping specifications.
The detectable area is a sphere of radius $r_d=0.15$.
The mapping specification is given by $c_p=(d, r_c, \theta, r_d)$, where $d\in[0.2,0.6]$, $r_c\in[0.0,0.3]$, and $\theta\in[-180^{\circ},180^{\circ}]$, corresponding to a cone with height $h$ and base radius $r_c$, whose apex is at the object’s center (Fig. \ref{fig:inspection_ex}(a)).
The robot's action space is continuous, controlling the end effector’s displacement.
Each episode lasts up to $500$ steps, and we trained all methods for 30 million steps.

\subsubsection{Task and curriculum settings}
Given the complexity of learning to manipulate an arm in a 3D environment, we have two curriculum levels with symbol numbers one and two.
To focus the evaluation on the complexity of task settings, we reduce the action space to three degrees of freedom corresponding only to the end-effector position.
The camera pose is determined by calculating the orientation at which the object appears in the center of the camera view by accessing the 3D coordinates of the next target symbol object.

\subsubsection{Results}
\begin{figure}[t]
    \begin{minipage}{1.0\hsize}
    \centering
    \includegraphics[width=85mm]{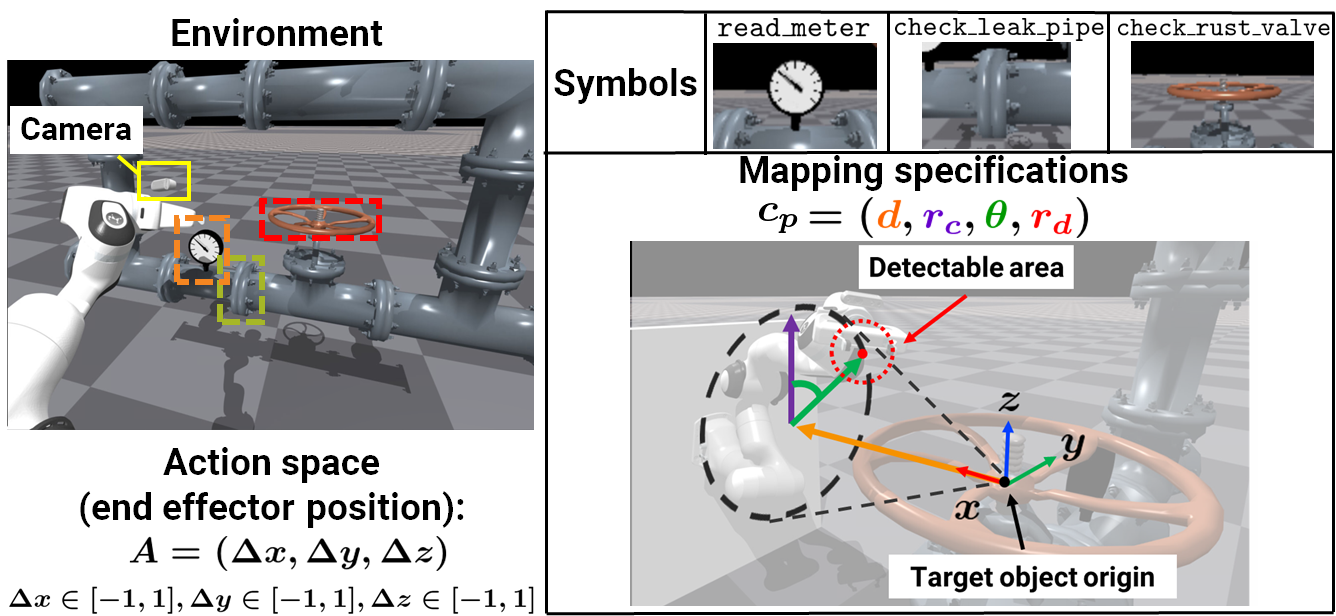}
    \subcaption{}
    \end{minipage}\\
    \begin{minipage}{1.0\hsize}
    \centering
    \includegraphics[width=85mm]{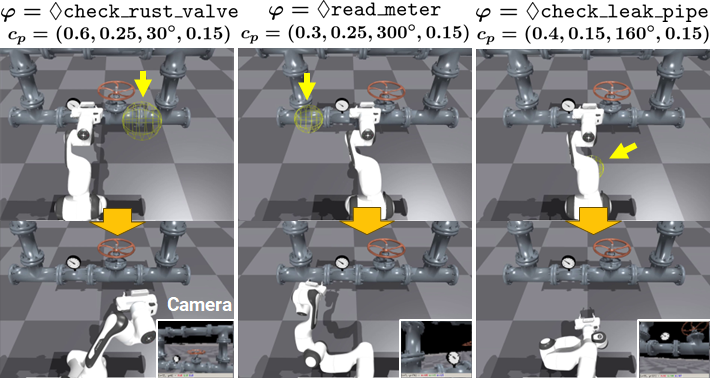}
    \vspace{-2mm}
    \subcaption{}
    \vspace{-1mm}
    \end{minipage}\\
    \caption{(a) Overview of environment of inspection task and specification settings: Three symbols are defined in environment, and mapping specifications are given by the 3D coordinates of each object’s center. (b) Examples of our trained policy behavior: Each row shows result of different LTL and mapping specifications. Top row images are at start of episode, where yellow spheres and arrows indicate positions of detectable area. the episode start, with yellow spheres and arrows indicating detectable areas. Bottom row images show the completed task, with the robot arm’s camera view at the bottom right of each image.}
    \label{fig:inspection_ex}
    \vspace{-2mm}
\end{figure}

\textbf{Performance evaluation against context-aware RL methods and curriculum strategies.}
Fig. \ref{fig:result_all} and Table \ref{tab:result} show the average success rate comparisons with each method for all the LTL tasks during training. 
The success rate of all the methods decreased due to the complexity of exploring a three-dimensional space with continuous action, despite fewer symbols in the LTL tasks than in the 2D navigation task.
While only ours and CARE advanced beyond the first curriculum level, ours completed every level and its average success rate was significantly higher than CARE’s.
These results show that our method most successfully learned the policy that follows an LTL task and its mapping specifications even in continuous action spaces in three-dimensional space, suggesting its applicability to the real world.
Fig. \ref{fig:inspection_ex}(b) shows examples of the behavior of our policy based on the instructed task and mapping specifications.
\subsection{Effects of Differences in Modulation Applied Layers}
Finally, we evaluated the impact of the number of modulations applied to each layer of the state encoder on the performance of \proposed.
Intuitively, the number of modulation adaptations to a layer can be seen as a parameter that controls the sensitivity of the mapping specification to conditioning on the state encoder and might affect the relationship between the state and the mapping specification.
Table \ref{tab:ablation} summarizes the results of the average success rate by changing the layers to which the modulation is applied for the three convolutional layers in Scenario 2 of the navigation and inspection tasks.
No significant differences were observed in the navigation task, where the dimensionality of the state and the mapping specification is low.
On the other hand, for the inspection tasks in a three-dimensional space, applying modulations to all layers improved the performance.
This suggests a correlation between the diversity of the mapping specifications and the sensitivity of the modulation for the state encoder.
Our method's applicability to various tasks can be enhanced by tuning this parameter based on the complexity of the state space and the symbol mapping specifications.

\section{Discussion}
Some challenges remain that require further attention as future work. 
First, the experiments in this paper assume that a symbol mapping specification explicitly constrains the state space, whereas defining a detailed specification for all the symbols is troublesome for humans.
Using such abstract instructions as natural language is a promising research direction; however, ambiguity in language representations may cause changes in mapping specifications for identical action sequences, and so the robot behavior must be quantified in the language representation \cite{spiegel2021guided}.
We showed that our method can successfully learn empirically under a limited number of symbols, but further evaluation is needed for scalability to LTL task sets with more symbols, including more diverse combinations of LTL operators.
Additionally, while we assume accurate symbol mapping and mapping specifications, addressing their uncertainties is required to enhance our method's applicability in the real world \cite{hatanaka2023reinforcement}.
\begin{table}[t]
    \centering
    \begin{tabular}{c|cc|c}
        \toprule
         \multirow{2}{*}{Methods}&  \multicolumn{2}{c|}{Navigation task}& \multirow{2}{*}{Inspection task}\\ \cmidrule{2-3}
          & Scenario 1&Scenario 2&\\  \midrule
         Soft-Module\cite{yang2020multi}&   0.65$\pm0.34^{*}$&0.34$\pm0.37^{**}$& 0.19$\pm0.34^{**}$\\
         CARE\cite{sodhani2021multi}&   0.88$\pm0.13^{*}$&0.79$\pm0.29^{*}$& 0.47$\pm0.21^{**}$\\
         Naive&   0.85$\pm0.19^{*}$&0.69$\pm0.34^{**}$& 0.17$\pm0.30^{**}$\\ 
         \proposed{} (Ours)&   \textbf{0.96}$\pm$0.05&\textbf{0.90}$\pm$0.16& \textbf{0.80}$\pm$0.09\\ \hline
         Anti-curriculum& 0.00$\pm0.00^{**}$& 0.00$\pm0.00^{**}$& 0.00$\pm 0.00^{**}$\\ 
         No curriculum& 0.018$\pm0.025^{**}$& 0.08$\pm0.17^{**}$& 0.20$\pm 0.06^{**}$\\ \bottomrule
    \end{tabular}
    \\\leftline{ *: $\textit{p}<.05$, **: $\textit{p}<.005$ by t-test.}
    \caption{Average success rate of testing policies in navigation and inspection task for all LTL tasks in curriculums: All results are averaged over three seeds with a rollout of 50 episodes of learned policies.}
    \label{tab:result}
\vspace{-2mm}
\end{table}
\begin{table}
    \centering
    \begin{tabular}{c|c|c}
    \toprule
         FiLM applied layer&  \begin{tabular}{c}Navigation task \\(Scenario 2)\end{tabular}& Inspection task\\ \midrule
         Only last layer& 0.88$\pm$0.20 & 0.71$\pm0.13^{**}$\\
         Last and 2nd layers&  \textbf{0.90}$\pm0.16$& 0.77$\pm0.13$\\
         All layers& 0.86$\pm0.22$ & \textbf{0.80}$\pm$0.09 \\ \bottomrule
    \end{tabular}
    \\\leftline{**: $\textit{p}<.005$ by t-test.}
    \caption{Average success rates for testing policies in navigation and inspection tasks across all LTL curricula when modulations are applied to different layers of a 3-layer CNN. These results use the best architectures identified in Sections \ref{navigation_task} and \ref{inspection_task}, applying modulations only to the last layer for navigation and to all layers for inspection.}
    \label{tab:ablation}
\vspace{-2mm}
\end{table}
\section{Conclusion}
We proposed a framework for \proposed{} that learns a policy for following diverse symbolic instructions with adjustable mapping specifications using LTL task representations.
Our experimental results show that our method performed the best for a navigation task with discrete action space and an inspection task with continuous action space in a simulated environment, even for long-horizon tasks.

\addtolength{\textheight}{-12cm}   


\bibliographystyle{IEEEtran}
\bibliography{main}

\end{document}